\documentclass[10pt,twocolumn,letterpaper]{article}

\usepackage[pagenumbers]{arxiv}      %

\newcommand{\appref}[1]{\ref{#1}}

\newcommand{\cmark}{\ding{51}}%
\newcommand{\xmark}{\ding{55}}%
\newcommand{\yes}{\color{Green}\cmark}
\newcommand{\no}{\color{red}\xmark}

\newcommand{\postcaption}{\vspace{-4pt}}
\newcommand{\pretablecaption}{\vspace{-8pt}}
\newcommand{\myparagraph}[1]{\vspace{0pt}\paragraph{#1}}
\newcommand{\mysection}[1]{\vspace{-5pt}\section{#1}\vspace{-4pt}}
\newcommand{\mysubsection}[1]{\vspace{-3pt}\subsection{#1}\vspace{-3pt}}

\def\name{\mbox{X-Capture}\xspace}

\definecolor{arxivblue}{rgb}{0.21,0.49,0.74}

\usepackage{multirow}
\usepackage{pifont}
\usepackage{array}
\usepackage{enumitem}
\usepackage[pagebackref,breaklinks,colorlinks,allcolors=arxivblue]{hyperref}

\title{\name:
An Open-Source Portable Device for Multi-Sensory Learning}

\author{Samuel Clarke \quad\quad
Suzannah Wistreich \quad\quad
Yanjie Ze \quad\quad
Jiajun Wu\vspace{1mm}\\
Stanford University
}

\begin{document}
\twocolumn[{
    \maketitle
    \begin{center}
        \captionsetup{type=figure}
        \vspace{-5mm}
        \includegraphics[width=\linewidth]{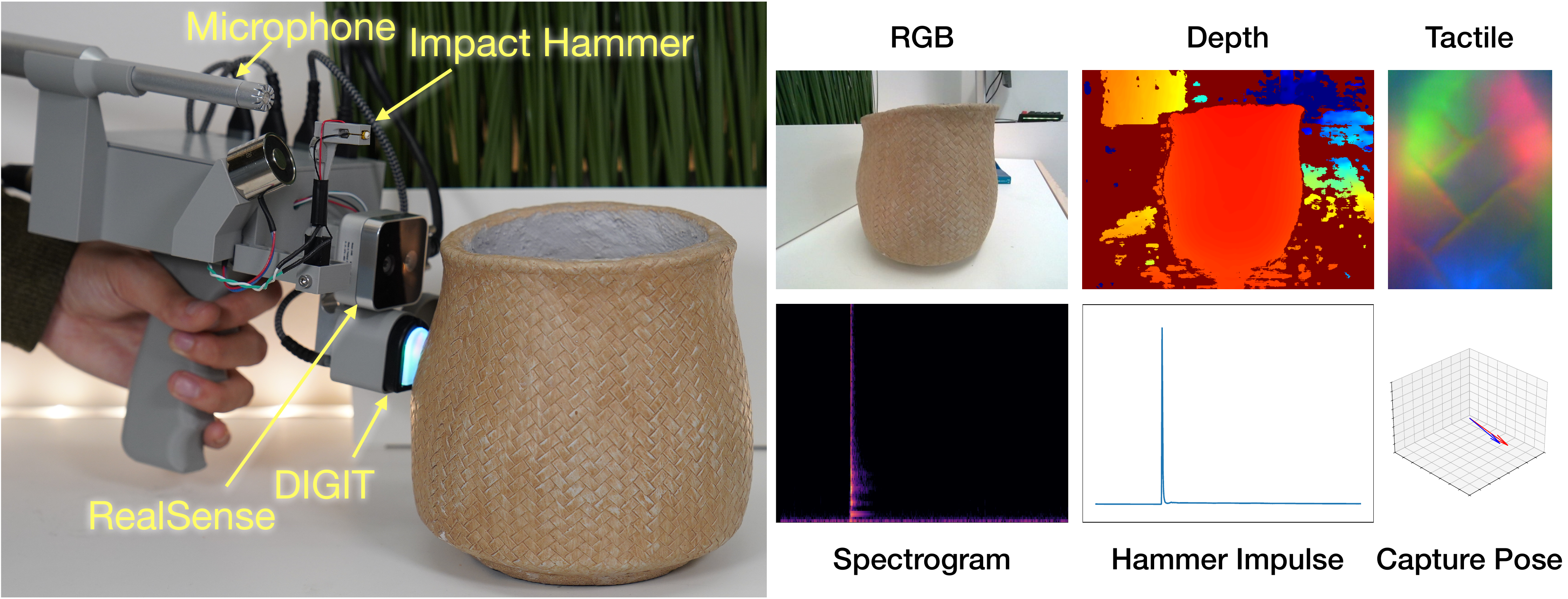}
        \vspace{-5mm}
        \captionof{figure}
        {{\bf \name for multi-sensory data capture}. (\textbf{Left}) The user captures tactile data from a vase in a living room. (\textbf{Right}) The sensor readings for each modality from the same probed point on the vase, as well as a visualization of the hammer impulse and 3D pose vectors for the image and tactile captures, shown in blue and red, respectively.}
        \label{fig:device_and_ui}
        \vspace{2mm}
    \end{center}    
}]

\begin{abstract}

Understanding objects through multiple sensory modalities is fundamental to human perception, enabling cross-sensory integration and richer comprehension. For AI and robotic systems to replicate this ability, access to diverse, high-quality multi-sensory data is critical. Existing datasets are often limited by their focus on controlled environments, simulated objects, or restricted modality pairings. We introduce \name, an open-source, portable, and cost-effective device for real-world multi-sensory data collection, capable of capturing correlated RGBD images, tactile readings, and impact audio. With a build cost under \$1,000, \name democratizes the creation of multi-sensory datasets, requiring only consumer-grade tools for assembly. Using X-Capture, we curate a sample dataset of 3,000 total points on 500 everyday objects from diverse, real-world environments, offering both richness and variety\footnote{Project page, hardware designs, and dataset are available at \mbox{\url{https://xcapture.github.io}}.}. Our experiments demonstrate the value of both the quantity and the sensory breadth of our data for both pretraining and fine-tuning multi-modal representations for object-centric tasks such as cross-sensory retrieval and reconstruction. \name lays the groundwork for advancing human-like sensory representations in AI, emphasizing scalability, accessibility, and real-world applicability.

\end{abstract}
    
\vspace{-10pt}
\mysection{Introduction}
As humans, we experience objects in our everyday environments through a combination of every sensory modality we possess. Each sensory modality provides us with unique information about the object, which can complement the information evident from other modalities. Touching what visually appears to be a ripe fruit can reveal that it is not in fact ripe yet or has a hidden bruise. Hearing a rigid drinking glass being tapped can disambiguate whether it is made of glass, crystal, or plastic. However, while each of these sensory modalities may complement each other, they are often highly correlated, as they each derive from the underlying physical properties of a given object~\cite{huh2024platonic}. Thus we are able to intuitively relate different modalities and generate expectations of other modalities from experiencing only one~\cite{johansson2009coding}.
In order for robots and agents to understand objects in the same way humans do, they must similarly possess such intuition about the relationships between objects' different sensory modalities.

In this paper, we focus on the sensory modalities of vision (both RGB and depth), sound, and touch---modalities for which popular commercially available sensors exist. Numerous powerful models have been developed to relate these modalities within a shared latent representation for interesting downstream cross-sensory inference and generation tasks~\cite{radford2021learning, girdhar2023imagebind, yang2024binding, xue2023ulip}. Such models rely on large datasets of examples correlating sensory modalities with each other. Although many such datasets exist, they suffer from key deficiencies that hamper their usefulness in training representations to enhance the understanding of real-world, in-the-wild objects. First, many multi-sensory datasets focus on \textit{scenes} rather than objects, reducing their relevance to applications where object understanding is a priority, such as robotic manipulation. Of those focusing on objects, many only include data collected from \textit{simulated} objects, or from real objects exclusively within controlled environments. Both simulated and controlled real data can present a large domain gap relative to data from real in-the-wild scenarios. Additionally, those collected in controlled environments often require expensive rigs and equipment, a drawback to both accessibility and scalability.
Finally, most object-centric multi-sensory datasets lack breadth in the sensory modalities they correlate, often linking only two sensory modalities, such as touch-vision or audio-vision. This hinders representations from forming \textit{direct} rather than emergent alignment among more than two modalities.

To address each of these deficiencies, we introduce \name, a portable, low-cost device for capturing \textit{correlated} RGBD images, tactile images, and impact audio samples from objects in the wild. Our device connects to and is powered by a laptop, with a user interface (UI) that visualizes data during the collection process. We ensure that each sensor takes independent readings (\textit{e.g.}, the touch sensor is not visible in camera images) through careful design of the device and UI, and we explicitly measure an input-output relationship for both touch and audio. These features address common shortcomings of many existing object-centric datasets and collection methods. We open-source the mechanical design and parts list of the device, with a total bill of materials of less than \$1000 at time of writing. The device requires only a consumer-grade 3D printer and soldering iron to assemble. Figure~\ref{fig:device_and_ui} shows the X-Capture device in use, along with sensor readings for each modality from a probed sample object. As shown in Table~\ref{tab:device_comparison}, our device is more versatile, portable, and relatively cheaper compared with other data-capturing devices. 

We collect a sample dataset of 500 different objects with our device and show how our dataset can be used to fine-tune existing multi-sensory representations to improve their performance in object-centric benchmark tasks such as cross-sensory retrieval, reconstruction, and detection.

\mysection{Related Work}
\label{sec:related}

Among large foundation models are many popular multi-sensory representation models, which attempt to acquire implicit knowledge of the physical world through their representations. Contrastive Language-Image Pretraining (CLIP) trained a representation between images and their text captions to relate images and text~\cite{radford2021learning}. ImageBind used CLIP as a backbone, along with separate datasets linking images to audio, depth, and thermal data, to train disparate modality-specific encoders to share a single multi-sensory representation~\cite{girdhar2023imagebind}. Though each dataset was used to train the model to link images with only one other modality, they showed evidence of \textit{emergent} alignment between other modalities. A Unified Representation of Language, Images, and Point Clouds (ULIP)~\cite{xue2023ulip} trained a unified representation of text descriptions, images, and point clouds of objects, showing that the representation was made measurably stronger by aligning all three modalities simultaneously during training, rather than only two at a time as ImageBind had done. Later works used both real and simulated tactile data to bind tactile images to the CLIP representation space as well~\cite{fu2024a, yang2024binding, cheng2024touch100k, yu2024octopi}.

\setlength{\tabcolsep}{1pt}
\begin{table}[t]
\centering
\begin{tabular}{lcccccc}
\toprule
\multirow{2}{*}{\textbf{Device/Setup}} & \multicolumn{4}{c}{\textbf{Object Properties}} & \multirow{2}{*}{\textbf{Port.}} & \multirow{2}{*}{\textbf{Cost}} \\
\cmidrule{2-5}
                 & \textbf{RGB} & \textbf{3D} & \textbf{Audio} & \textbf{Touch} & \\
\midrule
UBC ACME~\cite{pai2001scanning} & \yes & \yes & \yes & \yes & \no & N/A* \\
RealImpact~\cite{clarke2023realimpact} & \yes & \yes & \yes & \no & \no & \$8,000 \\
Obj.Folder Real~\cite{gao2023objectfolder} & \yes & \yes & \yes & \yes & \no & \$11,000 \\
TVL~\cite{fu2024a} & \yes & \no & \no & \yes & \yes & \$560 \\
\midrule
\textbf{\name (Ours)}   & \yes           & \yes            & \yes            & \yes              & \yes    & \$1000   \\
\bottomrule
\end{tabular}
\pretablecaption
\caption{Comparing prior multi-sensory object-centric data capture devices for what object properties they are designed to capture, their portability (``Port.''), and their estimated cost. Note that by ``Audio'' we are referring to impact sounds. The \name device captures images, point clouds, impact audio, and tactile data from objects in a portable and affordable package. (*UBC ACME~\cite{pai2001scanning} used equipment which is no longer commercially available at time of writing.)}
\label{tab:device_comparison}
\vspace{-4mm}
\postcaption
\end{table}
\setlength{\tabcolsep}{6pt}

\setlength{\tabcolsep}{1pt}
\begin{table}[t]
\centering
\begin{tabular}{lcccccc}
\toprule
\multirow{2}{*}{\textbf{Dataset}} & \multirow{2}{*}{\textbf{Obj.}} & \multicolumn{4}{c}{\textbf{Correlated Modalities}} & \multirow{2}{*}{\textbf{Env.}} \\
\cmidrule{3-6}
                 &                  & \textbf{RGB} & \textbf{3D} & \textbf{Audio} & \textbf{Touch} & \\
\midrule
Feeling of Success~\cite{calandra2017feeling}        & 106             & \yes         & \yes            & \no            & \yes               & T      \\
VisGel~\cite{li2019connecting}        & 195               & \yes          & \no            & \no             & \yes              & T     \\
Touch and Go~\cite{yang2022touch}        & 3971              & \yes           & \no            & \no            & \yes              & W      \\
SSVTP~\cite{kerr2022ssvtp}        & N/A              & \yes           & \no            & \no            & \yes              & T       \\
HCT~\cite{fu2024a}        & N/A              & \yes           & \no            & \no            & \yes              & W       \\
Greatest Hits~\cite{owens2016visually}        & N/A              & \yes           & \no            & \no            & \yes              & W      \\
RealImpact~\cite{clarke2023realimpact}        & 50            & \yes           & \yes            & \yes            & \no              & C       \\
ObjectFolder 2.0~\cite{gao2022objectfolder}        & 1000              & \yes           & \yes            & \yes            & \yes              & S       \\
ObjectFolder Real~\cite{gao2023objectfolder}        & 100              & \yes           & \yes            & \yes            & \yes              & C       \\
\midrule
\textbf{\name (Ours)}        & 500              & \yes           & \yes            & \yes            & \yes              & W       \\
\bottomrule
\end{tabular}
\pretablecaption
\caption{Comparing publicly available multi-sensory datasets by their number of objects, their \textit{correlated} sensory modalities, and the environments in which they are collected (C=Controlled, S=Simulation, T=Tabletop, and W=Wild). The \name dataset includes the widest breadth of correlated sensory modalities of a dataset collected in the wild.}
\label{tab:dataset_comparison}
\vspace{-5mm}
\postcaption
\end{table}
\setlength{\tabcolsep}{6pt}

Training these representations required a curated dataset linking two or more distinct sensory modalities. Many datasets link images with text through datasets of human-captioned images~\cite{schuhmann2021laion, schuhmann2022laion, gadre2024datacomp}.
Other datasets link sensory signals from two or more modalities, the majority of which consist of multi-sensory data from simulated environments~\cite{gao2021objectfolder, gan2020threedworld}, with some works showing how such data could be used to train models for downstream tasks in either simulated~\cite{hong2024multiply} or real~\cite{gao2022objectfolder, gao2023sonicverse} embodied environments. However, the ObjectFolder Benchmark showed that models fine-tuned with their proposed dataset of 100 \textit{real} objects generally performed significantly better on real-world embodied tasks than those trained only with simulated data~\cite{gao2023objectfolder}. Many multi-sensory datasets were collected using automated setups to scale up data collection~\cite{clarke2023realimpact, kerr2022ssvtp, pai2001scanning, calandra2017feeling, li2019connecting}, but similar to those of ObjectFolder Benchmark, such setups are often costly to replicate, collected in only controlled environments, and naturally impose some constraints on the types of objects which can be scanned. See Table~\ref{tab:device_comparison} for a comparison of \name to relevant data collection setups and devices, and Table~\ref{tab:dataset_comparison} for a comparison of our dataset to prior multi-sensory datasets. We include additional details in Appendix~\appref{app:dataset_device_comparison}.

Collecting multi-sensory data \textit{in situ} can produce datasets with distributions matching expected test conditions of embodied agents more closely. The authors of the Touch and Go dataset press a GelSight sensor against objects in the wild while holding a webcam behind the sensor~\cite{yang2022touch}. The authors of the Greatest Hits dataset~\cite{owens2016visually} similarly took video and audio samples of a drumstick striking objects. But without measuring inputs explicitly, such as the contact location and the pressing or striking force, it is impractical to infer the essential input-output relationship between force and touch or impact and sound of an object.

Hand-held sensorized devices can combine the best of both worlds, offering the inter-sensor measurement consistency of sophisticated setups with the portability needed to collect data in situ. Multiple works designed handheld devices for capturing correlated vision and tactile signals~\cite{fu2024a, dou2024tactile}. The authors of~\cite{bhattacharjee2016data} designed a handheld device for collecting multi-sensory tactile data to characterize thermal properties of objects. Numerous works have proposed sensorized handheld devices that approximated robot end-effectors~\cite{chi2024universal,young2021visual,song2020grasping}. Humans used these devices to collect demonstration data by performing tasks with everyday objects, which could be used to train a policy for a robot arm or mobile manipulator~\cite{shafiullah2023bringing,ha2024umi}. Our device brings the advantages of handheld data collection devices into a unified framework for measuring vision, touch, and audio data from objects, using commercially available sensors popular in many embodied learning applications.

\begin{figure}[t]
    \centering
    \includegraphics[width=\linewidth]{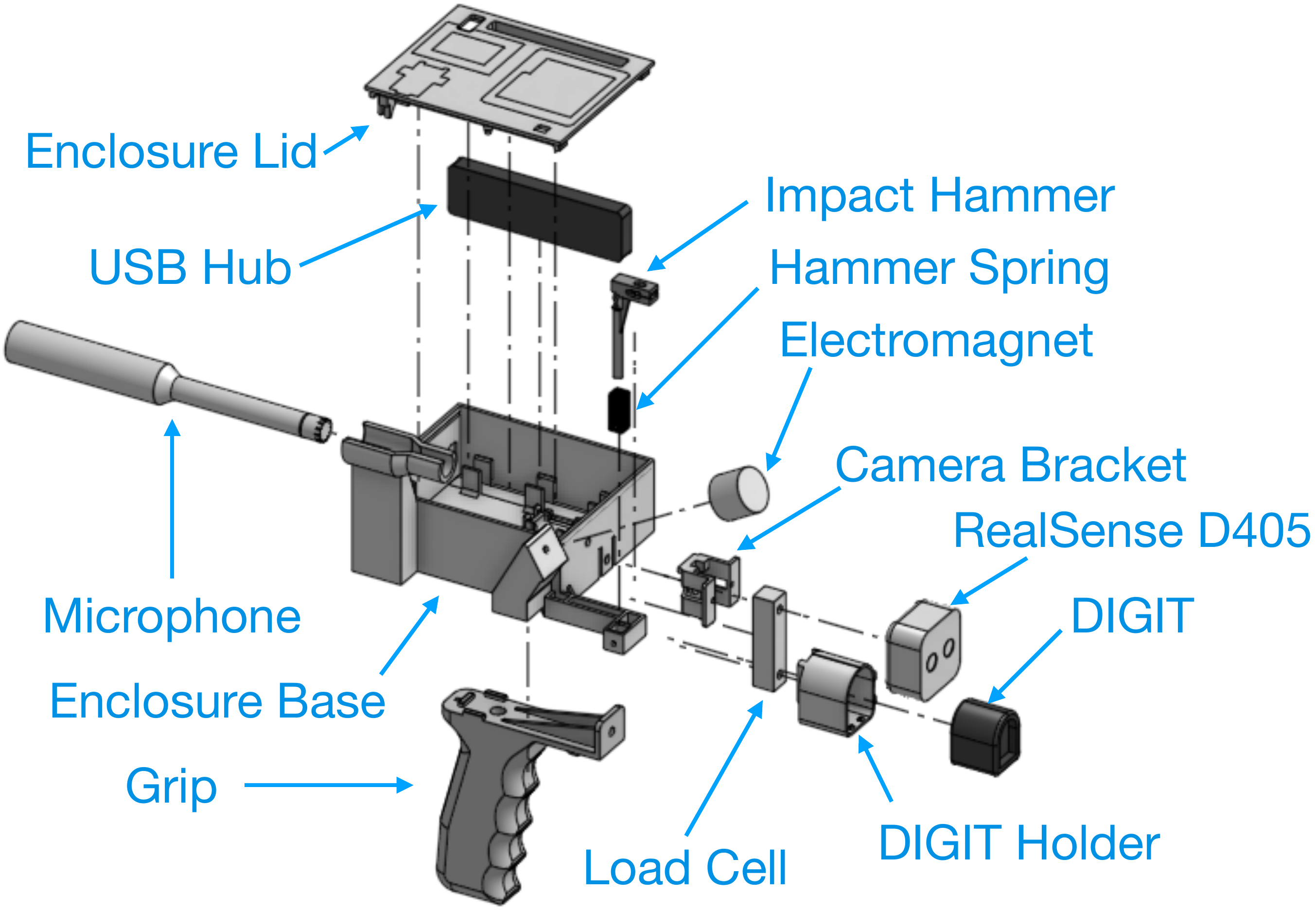}
    \caption{Exploded view of \name. The device rigidly constrains all sensor assemblies into fixed relative poses on a compact chassis with an ergonomic grip. Wires and circuitry are not shown.}
    \label{fig:exploded}
    \vspace{-0.2in}
\end{figure}

\mysection{The \name Device}
\label{sec:device}
The \name device supports sensing correlated RGB, depth, tactile, and impact audio samples using a combination of six distinct sensors. It is compact and power-efficient, making it as convenient as possible to carry, only requiring control and power from a laptop.
The chassis is constructed out of 3D printed PLA, with an ergonomic grip supporting an enclosure. The enclosure houses a USB hub that routes power and communications to all sensors and connects to a laptop with a single USB-C cable. Figure~\ref{fig:exploded} and the supplementary video show how the device and sensor assemblies physically fit together. We detail the compositions of the sensor assemblies and how they are used below, organized by sensory modality. See Appendix~\appref{app:hardware} for additional details on the hardware, including internal photos, a bill of materials, and support for alternative sensors, and the supplementary video for a sequential breakdown of how the sensors and assemblies fit together.

\mysubsection{Touch}

For tactile sensing, \name uses an assembly mounted on the front of the device consisting of a DIGIT sensor~\cite{lambeta2020digit} attached to a load cell. Vision-based tactile sensors are popular for both benchmark datasets~\cite{gao2023objectfolder} and robotic applications~\cite{kerr2022ssvtp, ai2024robopack, suresh2024neuralfeels}. We choose the DIGIT sensor for its commercial availability and relative durability~\cite{lambeta2020digit}. However, the images the sensor collects during contact with an object are not only a function of the object's intrinsic geometry and stiffness at the point of contact, but also \textit{extrinsic} factors such as the angle and the force at which the DIGIT sensor presses against the object. Thus, we design a novel assembly to explicitly measure this input-output relationship by annotating DIGIT tactile image readings in real time with relative angle and calibrated normal forces using an accelerometer and load cell, respectively. The device additionally uses the accelerometer to subtract the gravitational force exerted by the DIGIT sensor at the current device angle, dynamically re-zeroing the pressing force. This method uses low-cost, commercially available sensors to obviate the need to collect thousands of training points with a \$6,000 force sensor for estimating calibrated forces from the vision-based sensor~\cite{do2023densetact, helmut2024learning,yuan2017gelsight}. While collecting tactile data, our device automatically takes a snapshot of the DIGIT's tactile image at a pressing force within a 0.5N range of different target levels. We use 10, 15, and 20N as target levels during data collection.

\mysubsection{Vision and Depth}
\name captures RGBD images of objects with a RealSense D405 stereoscopic depth camera. The camera is compact and lightweight, with a 7cm minimum distance for measuring depth, lending itself well to close-up capture of objects. \name snapshots both the RGBD image and the accelerometer state simultaneously, such that the user can ensure a consistent angle of approach between the RGBD image and the tactile reading of a given point.

\mysubsection{Audio}
\name collects audio samples from objects by striking them with an impact hammer and recording the resulting impact sound with a microphone situated behind the point of impact.
The impact hammer is 3D printed from PLA, with a Thorlabs PK2JA2P1 piezo stack attached to a small steel rod at its tip. The piezo stack measures the impact force as it excites the object and produces the sound, in order to measure the input-output relation between the acting contact forces and resulting sound at each point. The base of the hammer fits into a 3D-printed elastomer base which functions as a spring. For each audio sample, the user pulls the hammer back until it adheres to an electromagnet mounted behind the hammer, storing potential energy in the elastomer base. After a pre-configured delay, the device automatically deactivates the electromagnet, releasing the hammer to strike the object in a silent and repeatable manner. The device records the impact sounds with a Dayton Audio EMM6 measurement microphone, which has a relatively uniform frequency response within the range of human-audible frequencies.

The impact hammer and the microphone's outputs are recorded by a HiFiBerry DAC2 ADC Pro board attached to a Raspberry Pi Zero 2W. The HiFiBerry board ensures that the hammer and microphone signal recordings are time-synchronized with minimal noise. The board supports high sample rates as well as digitally controlled gains, such that our device can dynamically and repeatably adjust the volume gain of each recording according to the loudness of each object's impact sounds. This ensures that the audio signal is recorded at a maximal level without clipping.

Finally, we design a custom PCB which provides precise power and input conditioning, as well as stable physical connections for all of the above electrical components. This novel assembly collects impact audio data at a comparably high fidelity to prior works~\cite{clarke2023realimpact, gao2023objectfolder}, while significantly optimizing power usage, size, and cost (\$130 vs. \$2,000), to improve portability and affordability.

\mysubsection{User Interface and Workflow}
\label{sec:ui_workflow}
Similar to the view shown at the right of Figure~\ref{fig:device_and_ui}, the UI visualizes all modalities of data in one screen to provide important feedback during data collection, allowing for retakes and comparisons to previous captures.  For each example, the user captures readings from each modality at one specific point on the object. The user first positions \name's RGBD camera such that a target point on the object is centered in the image, at a depth 8 to 13 cm away from the camera, as measured by depth image and displayed on a bar below the image. The UI displays the captured RGB and depth images, with the target point marked by small crosshairs superimposed at the center of both images. The user then presses the DIGIT sensor against the same point, using both the crosshairs on the RGB image and a display of the current angle of the device's accelerometer as a guide to ensure contacting the object at the same point and angle. During contact, a bar below the tactile image displays the current pressing force. The user pushes gradually up to 20N with snapshots automatically collected at 10, 15, and 20N. Finally, the user positions the impact hammer to hover over the target point, initiates a recording, pulls the hammer back to the electromagnet, then holds the device still while the electromagnet releases the hammer to strike the object after a delay. The UI displays the recorded spectrogram of the impact audio, as well as a time-domain graph of the impulse measured by the hammer, so the user can verify that the hammer made a single, clean impulse. See the supplementary video for a demonstration of this workflow.

\begin{figure}[t]
    \centering
    \includegraphics[width=\linewidth]{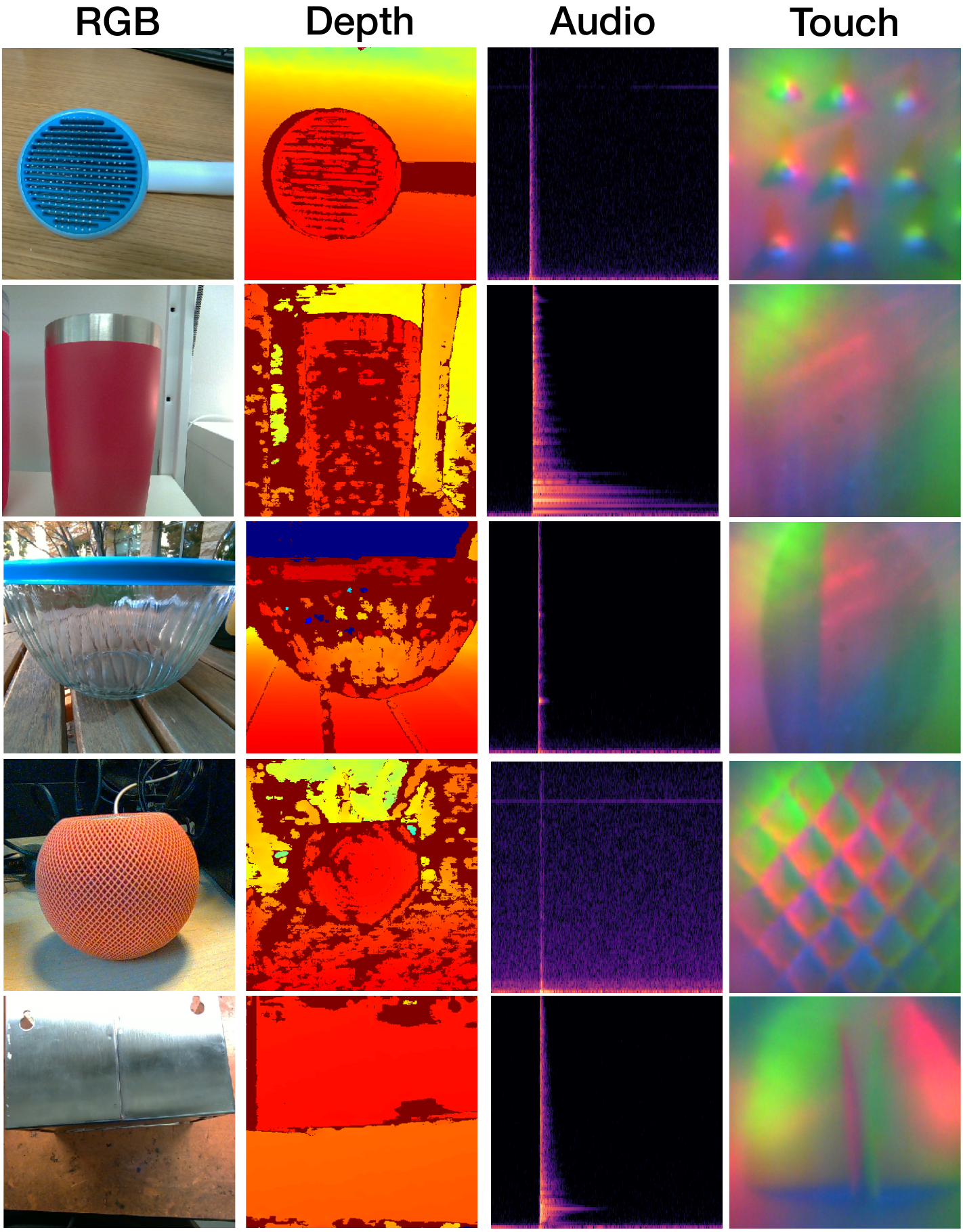}
    \vspace{-0.25in}
    \caption{Example multi-sensory data points from the \name dataset. Each row shows aligned multi-sensory data captured from a single point on an object in a distinct natural environment: (from left) RGB image centered on the point, depth image, impact audio spectrogram, and tactile image. (Objects, from top: Cat Brush, Insulated Steel Cup, Glass Storage Bowl, Computer Speaker, and Sheet Metal Container.)}
    \label{fig:montage}
    \vspace{-4mm}
    \postcaption
    
\end{figure}

\vspace{-7pt}
\mysection{The \name Dataset}
\vspace{5pt}

To evaluate our device's data collection pipeline, we present a novel dataset of correlated vision, tactile, audio, and post-processed point cloud data from a total of 3000 points on 500 objects in real-world environments, collected in just under three weeks. Our multi-sensory dataset includes data from objects across a diverse class of materials, geometries, and functional uses. Aided by our device's flexible capabilities and portable design, we capture a wide breadth of objects encountered in everyday settings. We include further details of the objects and environments comprising our dataset in Appendix~\appref{app:dataset_details} and a comprehensive comparison of our datasets to existing alternatives in Appendix~\appref{app:dataset_device_comparison}.

\mysubsection{Collection Procedure}
We collect data across nine different environments-- one indoor workspace with relatively controlled conditions, and eight in-the-wild locations, ranging from everyday indoor settings to a dynamic outdoor area. We capture 300 diverse objects in the indoor workspace, a quiet office with consistent lighting, providing a distribution baseline. We probe the remaining 200 objects in diverse natural environments, including a Kitchen, Bathroom, Home Office, Workshop, Bedroom, Laundry Room, Living Room, and an outdoor Picnic Table. We capture an equal number of objects in each in-the-wild environment.

For each object, we collect readings of each modality from six distinct points on the object. We choose point locations which cover the breadth of each object's surface and capture unique local features. Each of the six points has a corresponding RGBD image, impact audio recording, and tactile impressions captured at 10N, 15N, and 20N. We use our device and UI to manually register these sensory readings with each other at each point by following the procedure described in Section~\ref{sec:ui_workflow}.

\mysubsection{Dataset Labeling and Post-Processing}
We provide a brief text description of each object, noting salient materials and the relevant state of the object (\eg, if a can is full or empty). We further postprocess our data, using the RGB and depth data to estimate point clouds of objects and using the noted recording gains and hammer signals to normalize the audio. We include details of both these post-processing steps in Appendix~\appref{app:dataset_postprocessing}.

\begin{figure*}[t!]
    \centering
\includegraphics[width=1.0\linewidth]{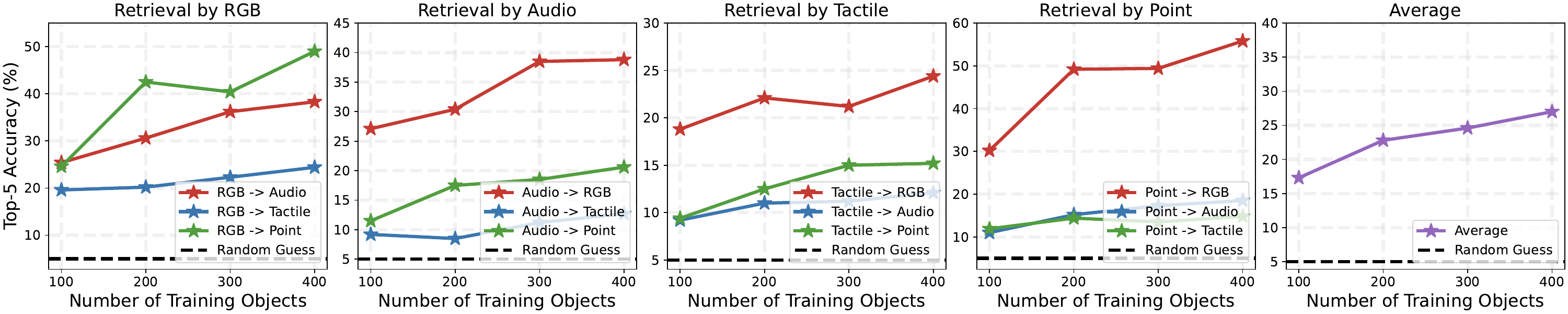}
\vspace{-20pt}
    \caption{Comparing test retrieval performance of our cross-modal encoders trained with varying quantities of objects, with each plot grouping results by the query modality used for retrieval. The right-most plot shows an average across all modality combinations.}
    \label{fig:scaling law}
    \postcaption
    \vspace{-3pt}
\end{figure*}

\mysection{Experiments}
We validate the usefulness of the sample dataset we collect with the \name device with three popular multi-sensory benchmark tasks: cross-sensory retrieval, image generation, and point cloud generation. We also demonstrate how we can use our data to train an audio-based object detector. Unless otherwise noted, we randomly split our dataset into 400 training and 100 test objects. See Appendix~\appref{app:training_details} for additional details on training and testing procedures.

\mysubsection{Baselines}
We evaluate recent cross-sensory representation frameworks on our dataset. The ImageBind framework~\cite{girdhar2023imagebind} provides encoders for the RGB, audio, and depth modalities, each pretrained on  web-scale datasets pairing images with each modality, but the framework lacks encoders for point clouds or tactile readings. In order to cover all of the modalities we provide in our dataset, we also evaluate the performance of a combination of pretrained encoders from different sources, which each specialize in a specific modality or cross-sensory representation. We encode RGB images with CLIP's~\cite{radford2021learning} pretrained ViT-L encoder. For tactile images, we use the ViT-L encoder publicly released with~\cite{fu2024a}, which has been pretrained on the TVL dataset. We encode our audio recordings with the Audio Spectrogram Transformer (AST)~\cite{gong2021ast}, with publicly released weights from pretraining on ImageNet~\cite{deng2009imagenet} and AudioSet~\cite{gemmeke2017audio}. Finally, for point clouds, we use the PointBERT model~\cite{yu2022point} with publicly-released weights from pretraining on ULIP~\cite{xue2023ulip}.

For both the ImageBind encoders and the ensemble of modality-specific encoders, we evaluate three different training configurations. We first evaluate the publicly-released out-of-the-box pretrained weights. Then we test fine-tuning all pretrained models on our dataset's training set, using two distinct formulations of the contrastive InfoNCE loss~\cite{oord2018representation}. In the ``Image Loss'' formulation, we fine-tune according to a symmetric InfoNCE loss between images and each other available modality, the same technique used by ImageBind. In the ``Cross-Sensory Loss'' formulation, we fine-tune according to a symmetric InfoNCE loss between each pairing of \textit{all} modalities, similar to the loss used among vision, language, and point cloud encoders in ULIP. For all configurations, we keep the encoders for the RGB image modality frozen during training.

\begin{table}[t!]
\centering
\setlength{\tabcolsep}{2pt}
\resizebox{0.48\textwidth}{!}{
\begin{tabular}{lccccccc}
\toprule 

ImageBind  & \multicolumn{2}{c}{RGB$\rightarrow$} & \multicolumn{2}{c}{Audio$\rightarrow$} & \multicolumn{2}{c}{Depth$\rightarrow$} & \multirow{2}{*}{Avg.}  \\

\cmidrule(lr){2-3}\cmidrule(lr){4-5}\cmidrule(lr){6-7}
 Top-5 Accuracy (\%)  & Audio & Depth & RGB & Depth & RGB & Audio  \\
\midrule

Random Guess & 5.0& 5.0& 5.0& 5.0& 5.0& 5.0& 5.0\\
  Out-of-the-Box Pretrained &  6.2 & 9.2 & 6.0 & 4.6 & 8.0 & 6.8 & 6.8\\

Fine-Tune w/ Image Loss & \textbf{39.0} & 62.4 & 37.4 & 20.2 & 63.2 & 22.6 & 40.8\\

Fine-Tune w/ Cross-Sens. Loss & 38.0 & \textbf{64.8} & \textbf{41.0} & \textbf{21.2} & \textbf{64.4} & \textbf{24.4} & \textbf{42.3} \\

\bottomrule
\end{tabular}
}
\pretablecaption
\caption{Cross-sensory retrieval top-5 accuracies of ImageBind trained with different strategies using our dataset. The top and bottom column headers denote the query and retrieved modalities, respectively. Out-of-the-Box weights do not generalize well to our data, whereas Fine-Tuning with a Cross-Sensory Loss outperforms other configurations, across all modalities.}
\label{tab:imagebind}
\vspace{-1mm}
\postcaption
\end{table}

\begin{table}[t!]
\centering
\setlength{\tabcolsep}{2pt}
\resizebox{0.48\textwidth}{!}{
\begin{tabular}{lccccccc}
\toprule 

ImageBind  & \multicolumn{2}{c}{RGB$\rightarrow$} & \multicolumn{2}{c}{Audio$\rightarrow$} & \multicolumn{2}{c}{Depth$\rightarrow$} & \multirow{2}{*}{Avg.}  \\

\cmidrule(lr){2-3}\cmidrule(lr){4-5}\cmidrule(lr){6-7}
 Top-1 Accuracy (\%)  & Audio & Depth & RGB & Depth & RGB & Audio  \\
\midrule
Random Guess  & 16.7& 16.7& 16.7& 16.7& 16.7& 16.7& 16.7\\
  Out-of-the-Box Pretrained & 16.3 & 20.0 & 16.0 & 18.5 & 19.3 & 18.2 & 18.1\\

Fine-Tune w/ Image Loss & 20.3 & 32.3 & 19.8 & 20.0 & \textbf{45.8} & 20.7 & 26.5\\

Fine-Tune w/ Cross-Sens. Loss & \textbf{24.0} & \textbf{34.5} & \textbf{20.3} & \textbf{22.0} & 44.7 & \textbf{22.3} & \textbf{28.0} \\

\bottomrule
\end{tabular}
}
\pretablecaption
\caption{Contact localization top-1 accuracies of ImageBind trained with different strategies using our dataset. The top and bottom column headers denote the query and retrieved modalities, respectively. Out-of-the-Box weights generalize poorly to our data, and Fine-Tuning with a Cross-Sensory Loss achieves best average performance.}
\label{tab:imagebind intra}
\vspace{-4mm}
\postcaption
\end{table}

\mysubsection{Cross-Sensory Retrieval}
\label{sec:cross-sensory_retrieval}
Cross-sensory retrieval assesses models' abilities to connect multi-sensory information, akin to the human ability to intuitively link sight, sound, and touch, to make valuable inferences about unknown properties of objects. Consequently, it has become a common benchmark in cross-sensory learning \cite{gao2023objectfolder, girdhar2023imagebind,yang2024binding}. We formulate the task as an inter-object classification task, testing each cross-sensory frameworks' performance as follows: given a randomly selected point from each of $N$ objects, where $N=100$ for our test dataset, can the framework correctly associate a sensory reading from one modality with another from the same point?

 We show the top-5 accuracies for ImageBind in Table~\ref{tab:imagebind}. Note that for 100 objects, the expected value of random selection would be 5\% under our test conditions. While the encoders have been pretrained on very diverse datasets, their out-of-the-box weights struggle on our object-centric data, especially with the audio modality. Our results also suggest that using a full cross-sensory loss comparing all modalities directly provides a generally stronger representation on our data. We show the results for the cross-sensory encoder ensemble in Table~\appref{tab:cross_modal} of Appendix~\appref{app:additional_retrieval}. Interestingly, in these results we see less of a clear performance advantage to training with the Cross-Sensory Loss versus the Image Loss, as compared to the advantage we observe in our ImageBind experiments. For modality pairings the ensemble and ImageBind have in common, such as RGB$\rightarrow$Audio, we see similar results to those of ImageBind.

\mysubsection{Cross-Sensory Contact Point Localization}

\label{sec:cross-sensory_contact}
Contact point localization assesses models' abilities to differentiate between sensory signals coming from different points on the \textit{same} object.
We thus formulate contact point localization as a classification task similar to cross-sensory retrieval, except that instead of comparing different sensory signals from a single point from one object to points of other objects, we compare different sensory signals from a single point to those of another modality from all $M$ points on the same object, where $M=6$ in our dataset.

We show the top-1 accuracies for ImageBind in Table~\ref{tab:imagebind intra}. In this task, the expected value of random selection is $\sim16.7\%$. Once again, ImageBind's pretrained weights struggle to outperform random chance on our object-centric data. However, this is clearly a difficult task even after fine-tuning. ImageBind seems to excel at differentiating object points with RGB and depth much more than it does with audio, perhaps because the impact sounds a single object produces at different points often have very minute differences. We see similar results for our cross-sensory encoder ensemble in Table~\appref{tab:cross_modal_intra} of Appendix~\appref{app:additional_localization}, where the association between RGB and point cloud is much stronger than associations between other modalities. 

\begin{figure*}[ht!]
    \centering
    \includegraphics[width=\linewidth]{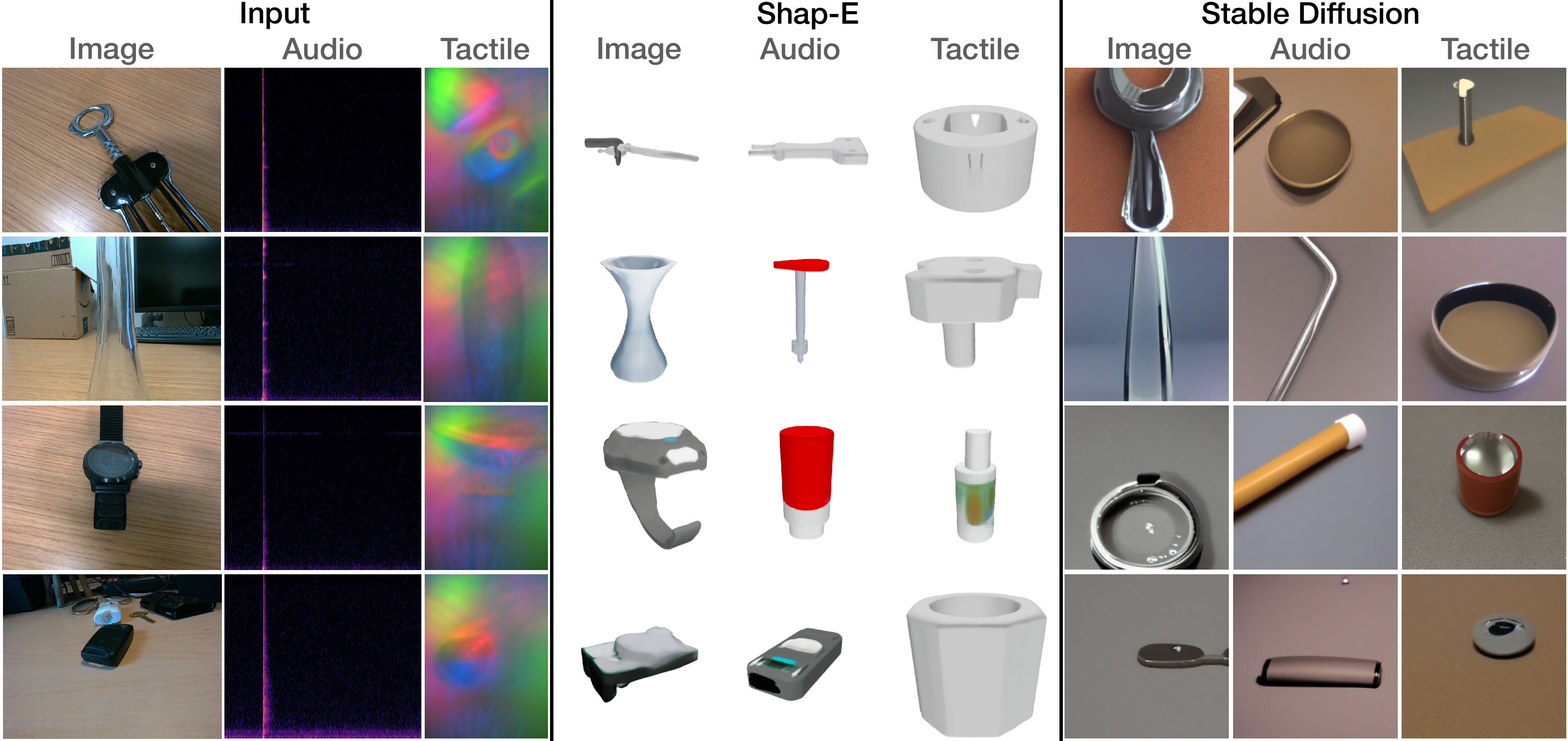}
    \caption{Results of using Shap-E~\cite{jun2023shap_e} to generate 3D neural radiance fields and Stable Diffusion~\citep{rombach2022stable_diffusion} to generate images from outputs of multimodal encoders which have been trained on our data to align to CLIP features. The three left columns show the RGB images, audio spectrograms, and tactile images inputted to their respective encoders. The next three columns show the neural radiance fields generated from using the outputs of the encoders from these RGB, audio, and tactile inputs, respectively, as input to Shap-E. The last three columns similarly show the images generated from using the outputs of the RGB, audio, and tactile inputs, respectively, as input to Stable Diffusion.
    }
    \postcaption
    \label{fig:generations}
\end{figure*}

\mysubsection{Scaling Law}
We claim that the \name device makes multi-sensory object-centric data collection more efficient than prior alternatives, allowing us to scale up the collection of a valuable dataset. Though we have already collected a sample dataset of 500 objects, we evaluate how a model's performance improves according to the quantity of our training objects we use during fine-tuning. The results in Figure~\ref{fig:scaling law} demonstrate that each modality continues to benefit from additional training examples, and the performance improvement seems far from plateauing at 400 objects, confirming the value of scaling up the \name dataset even further.

\mysubsection{Using \name Data as a Pretraining Set}
\label{sec:objectfolder_pretraining}
Multi-sensory object understanding is important in many different downstream applications, including robotic manipulation, anomaly detection, and generation. However, each potential application may involve unique data with a domain gap differentiating it from the data in existing large-scale datasets. Fortunately, such large-scale datasets can still be used to pretrain models, which can be fine-tuned on additional data for downstream applications. We evaluate whether data from \name could be similarly helpful by using it as a pretraining set.

We evaluate cross-sensory retrieval on the real object data from the ObjectFolder Benchmark~\cite{gao2023objectfolder}. Though ObjectFolder's subject matter is quite similar to ours, there is a significant domain gap in how each modality is recorded. Their RGB images are background-less renderings from 3D scans of objects rather than real images. Their tactile data is from a GelSight sensor, which has a different texture detail and lighting conditions than the DIGIT. Finally, they collect impact sounds in an acoustically-treated room, with objects suspended by string to reduce the contact-damping usually affecting objects' impact sounds in real-world settings.

We evaluate the cross-sensory encoder ensemble's vision, audio, and tactile encoders trained with three different configurations using the cross-sensory loss. In ``Our Pretrained Only'' configuration, we train encoders with only \name data. In ``Fine-Tuned Only'', we train the encoders only with data from the 70 real training objects from ObjectFolder. And in ``Our Pretrained + Fine-Tuned'', we pretrain the encoders with our data, then fine-tune on the real training objects from ObjectFolder. We evaluate cross-sensory retrieval on the 30 real test objects of ObjectFolder Benchmark and show the top-5 accuracies in Table~\ref{tab:pretrain and finetune}. The expected value of random selection is $16.7\%$. We see that ``Our Pretrained Only'' model outperforms random chance, but still performs rather poorly in this zero-shot generalization, likely struggling to surmount the domain gap without fine-tuning on ObjectFolder's data. However, ``Our Pretrained + Fine-Tuned'' model outperforms the ``Fine-Tune Only'' model, suggesting that pretraining with the \name dataset helps bridge the generalization gap in the low-data regime of the ObjectFolder Real training set. The improvement is evident in the audio modality, but not in the tactile modality, suggesting that the domain gap between different tactile sensors may be especially challenging.

\mysubsection{X-to-2D/3D Generation}
\label{sec:cross-generation}
For humans, the sound or feel of an object can often conjure a mental image of the object. We thus evaluate the ability of our representations aligned to CLIP image features to provide a useful representation for both a CLIP-based 3D implicit function generator and CLIP-based image generator. We use our dataset to train our tactile and audio encoders to align to the outputs of CLIP's ViT-L encoder of the corresponding image of each point. For this experiment, we use a mean-squared error (MSE) rather than contrastive loss to prioritize alignment to CLIP over cross-sensory association and differentiation. After training alignment on our training objects, we use the signals encoded from our held-out test objects as input to pretrained Shap-E~\citep{jun2023shap_e} to generate 3D implicit functions and to pretrained Stable Diffusion~\citep{rombach2022stable_diffusion} to generate images. Both models have been pretrained by their authors to use CLIP ViT-L features encoded from text to generate 3D implicit functions and images, respectively.

\begin{table}[t]
\centering
\setlength{\tabcolsep}{2pt}
\resizebox{0.48\textwidth}{!}{
\begin{tabular}{lccccccc}
\toprule 

  ObjectFolder-Real (30 Objects)& \multicolumn{2}{c}{RGB$\rightarrow$} & \multicolumn{2}{c}{Audio$\rightarrow$} & \multicolumn{2}{c}{Tactile$\rightarrow$} & \multirow{2}{*}{Avg.} \\
\cmidrule(lr){2-3}\cmidrule(lr){4-5}\cmidrule(lr){6-7}
 Top-5 Accuracy (\%) & Audio & Tactile & RGB & Tactile & RGB & Audio  \\
\midrule

Random Guess & 16.7 & 16.7 & 16.7 & 16.7 & 16.7 & 16.7 & 16.7 \\

Our Pretrained Only  &  26.7 & 18.3 & 25.0 & 19.2 & 14.2 & 22.5 & 21.0 \\

Fine-Tuned Only  & 25.8 & \textbf{30.0} & 24.2 & 29.2 & \textbf{32.5} & 25.8 & 27.9\\
Our Pretrained + Fine-Tuned & \textbf{38.3} & 28.3 & \textbf{36.7} & \textbf{30.0} & 27.5 & \textbf{30.0} & \textbf{31.8}\\
\bottomrule
\end{tabular}
}
\pretablecaption
\caption{Results of ablating both pretraining with our dataset and fine tuning with train objects from the ObjectFolder Real dataset~\cite{gao2023objectfolder} for downstream evaluation of cross-sensory retrieval on the real test objects from ObjectFolder.}
\label{tab:pretrain and finetune}
\postcaption
\vspace{-5mm}
\end{table}

We show qualitative results for both image generation and 3D neural radiance field generation in Figure~\ref{fig:generations}. The results vary substantially in quality, but they seem to provide some interesting insights into what features our encoders are able to glean from each respective sensory modality. We see that both shape and image generations from the image encoding tend to match the original object in both color and semantic features. Generations from audio encodings often successfully match the salient materials of the object, and in some cases, they show surprising semblance of unique geometric features of the object, such as in the cases of Shap-E's renderings from the audios of the bottle opener and the car key fob. Generations from tactile encodings seem to excel at matching the geometric features local to the contact point, such as the local curvature, and also occasionally match in material properties such as hardness. These results lend further evidence of the complementary information which can be inferred from different sensory modalities when interacting with objects.

\begin{figure}[t]
    \centering
    \includegraphics[width=\linewidth]{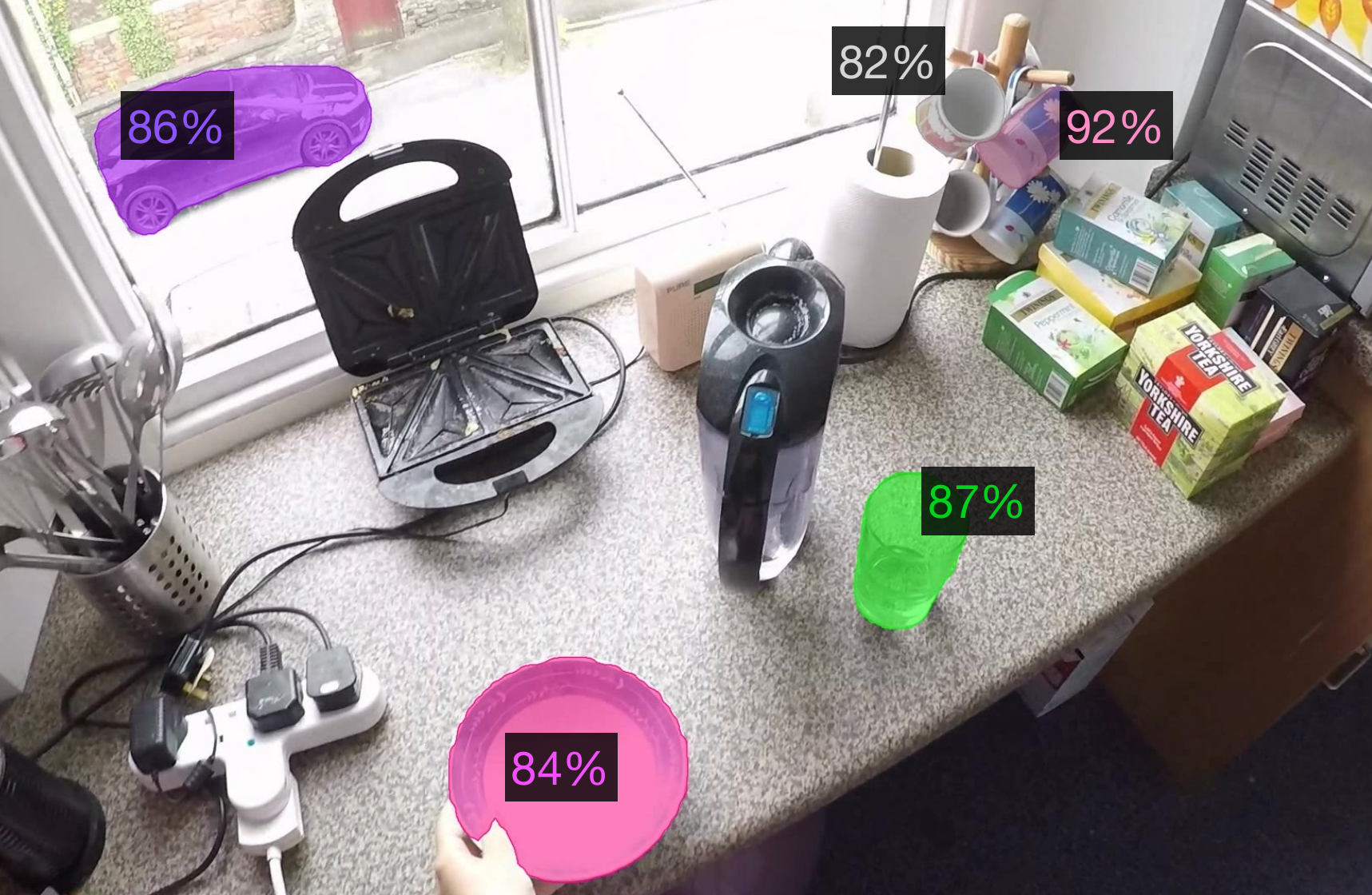}
    \vspace{-0.27in}
    \caption{Still frame from prompting Detic~\cite{zhou2022detecting} with our pretrained audio encoder's embedding of the real sound of the ceramic plate (at bottom, highlighted pink) being placed on the counter. Detic identifies the plate, as well as the ceramic mugs, drinking glass, and car, as likely sources of the sound. It successfully ignores appliances, cardboard tea boxes, and metal utensils.}
    \label{fig:epic_kitchens}
    \postcaption
    \vspace{-4mm}
\end{figure}

\mysubsection{Zero-Shot Audio-Based Object Detection}
\label{sec:audio_detection}
Similar to ImageBind~\cite{girdhar2023imagebind}, we use our dataset to train a CLIP-aligned audio embedding which can replace the input for the text-based Detic detection model~\cite{zhou2022detecting}. We train our audio encoder on our dataset contrastively to ViT/B-32 CLIP features, then use embeddings from this encoder to prompt Detic's CLIP-based object detector.
Though our dataset is collected with objects in the wild, all objects are captured in static configurations, whereas humans and robots often perceive objects through dynamic interactions. In order to test generalization to audio from such natural dynamic interactions, we use clips from the EPIC-KITCHENS-100 dataset~\cite{Damen2022RESCALING} of humans using their kitchens naturally, and select clips where a human impacts an object against a table or another object and use the audio embedding of the impact sound to prompt the Detic model. We show one such result in Figure~\ref{fig:epic_kitchens} and additional results in Figure~\appref{fig:additional_detection} of Appendix~\appref{app:additional_detection}. We include the original video clips with audio in our supplementary video. The detector mostly selects either the correct item or items of similar materials and acoustic properties. ImageBind does not provide weights from this task for comparison.

\mysection{Conclusion and Limitations}
We introduced \name, an open-source and low-cost device for collecting multi-sensory data in the wild. Using \name, we collected a sample dataset of correlated RGB, depth, audio, and tactile readings of 3,000 points from 500 objects in natural environments, enabling direct benchmarking of cross-sensory encoding frameworks and loss functions on retrieval and contact localization tasks. Our results suggest that cross-sensory representations can be strengthened by learning from object-centric data correlating as many sensory modalities as possible, and that pretraining on this data yields valuable representations that can be fine-tuned to improve performance on other object-centric tasks. However, a limitation of \name is that it captures objects in static configurations of environments, whereas humans and robots learn about objects interactively and dynamically while manipulating them. We hope our work inspires new, perhaps automated, collection efforts to further scale up multi-sensory learning from real objects.

\myparagraph{Acknowledgements.} We thank Roger Clarke, Ryan Williams, Anirudh Jain, Mark Rau, and Fernando Lopez-Lezcano for advice with the hardware design, Klemen Kotar, Le Xue, Stephen Tian, and Weiyu Liu for valuable conceptual discussions, and Andrej Krevl and Matt Wright for their facility support. This work is in part supported by NSF CCRI \#2120095 and RI \#2338203 and ONR MURI N00014-22-1-2740. S. Clarke is supported by the Meta PhD Fellowship.

{
    \small
    \bibliographystyle{ieeenat_fullname}
    \bibliography{abbreviations, cite}
}

\clearpage
\setcounter{section}{0}
\setcounter{table}{5}
\setcounter{figure}{6}
\renewcommand{\thesection}{\Alph{section}}
\maketitlesupplementary

\noindent The supplementary materials consist of:
\begin{enumerate}[A., leftmargin=20pt]
   \item A video showing the design and usage of the \name device
    \item A comprehensive comparison of the \name dataset to prior multi-modal datasets and devices
    \item Additional details on the device hardware: photos of the internals, a bill of materials, and information on support for alternative sensors
    \item Additional information on the \name dataset: more details on the objects and environments, as well as the postprocessing steps
    \item Additional experimental results and details
\end{enumerate}

\begin{figure}
    \centering
    \includegraphics[width=\columnwidth]{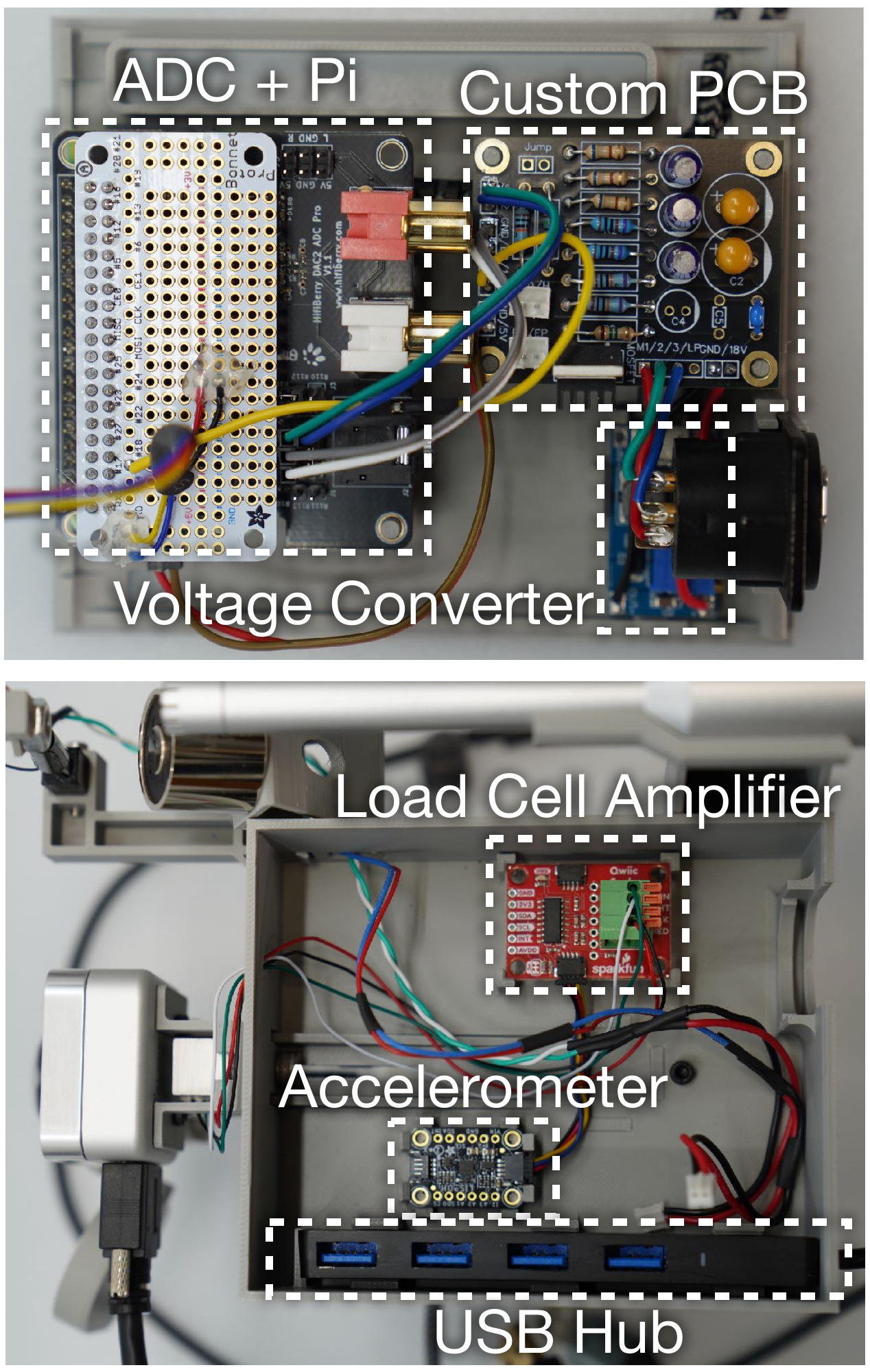}
    \caption{Photos of the \name device internals. (\textbf{Top}) The lid of the device has a white prototyping board stacked on top of a HiFiBerry DAC2 ADC Pro, which is also stacked on top of a Raspberry Pi Zero 2W single-board computer. Our custom PCB has circuitry for powering and filtering both the impact hammer and the microphone. The voltage converter, partially occluded in this photo by the microphone jack, provides 18V power for the microphone from USB-powered 5V. (\textbf{Bottom}) The base of the device houses an amplifier for the load cell signal, as well as an accelerometer. The USB hub provides USB A ports for the RealSense D405, DIGIT, Raspberry Pi, and voltage converter, connecting them all through a USB C connection to a laptop or desktop computer.}
    \label{fig:device_internals}
    \vspace{-0.01in}
\end{figure}

\section{Supplementary Video}
\phantomsection
\label{app:video}
The included video (\texttt{XCaptureVideo.mp4}) shows a breakdown of the layout and design of the \name device, then shows the process of using the device to collect data from an object, as well as the feedback the user receives from each sensor on the user interface during this process. We also include qualitative examples, with audio, from the generation and audio-based detection experiments described in Section~\appref{sec:cross-generation}~and~\appref{sec:audio_detection}, respectively.

Specific portions of the video referenced in the text begin at the following timestamps:
\begin{itemize}[leftmargin=20pt]
    \item \textbf{$[$00:49$]$} A breakdown of the hardware assemblies described in Section~\appref{sec:device}
    \item \textbf{$[$02:40$]$} A demonstration of using the device and user interface to collect data through the workflow described in Section~\appref{sec:ui_workflow}
    \item \textbf{$[$06:36$]$} Example video clips, with sound, from the audio-based detection experiment described in Section~\appref{sec:audio_detection}
\end{itemize}

\section{Comparison to Prior Multi-Sensory Datasets and Devices}
\phantomsection
\label{app:dataset_device_comparison}
While there are prior object-centric multi-sensory datasets, our dataset is the first of its kind to correlate RGB, depth, impact audio, and tactile sensing at a point-level of objects in the wild. This is made possible by the design of the \name device integrating both existing and novel sensor assemblies of difference sensory modalities into a single portable device. We compare to relevant prior datasets and prior devices in more detail below.

\paragraph{Prior Object-Centric Multi-Sensory Datasets}
We compare the \name dataset to prior works across three dimensions: the quantity of data, the sensory modalities included, and the data collection environment in Table~\appref{tab:dataset_comparison} of Section~\appref{sec:related}. Many datasets correlate only two or three sensory modalities. While some datasets also provide correlated RGB, depth, audio, and tactile data, the \name device supports capturing all of these modalities in a correlated fashion \textit{in the wild}.

With the exception of SSVTP~\cite{kerr2022ssvtp} and ObjectFolder~\cite{gao2022objectfolder, gao2023objectfolder}, these datasets do not explicitly correlate at the \textit{point} level of an object. For example the Feeling of Success~\cite{calandra2017feeling}, Touch and Go~\cite{yang2022touch}, and HCT~\cite{fu2024a} correlate tactile images with RGB images where the contact region is specifically occluded by the sensor. Greatest Hits~\cite{owens2016visually} includes videos of a wooden drumstick striking objects and surfaces, where exact contact location is not obvious from the videos due to motion blur at 30 frames per second. We use the \name device to manually register a reading of all four modalities to the same point, such that we can use our data to learn object-centric multi-sensory representations at a point-level resolution.

\begin{figure*}[h]
    \centering
    \includegraphics[width=\textwidth, height=1.05\textheight, keepaspectratio]{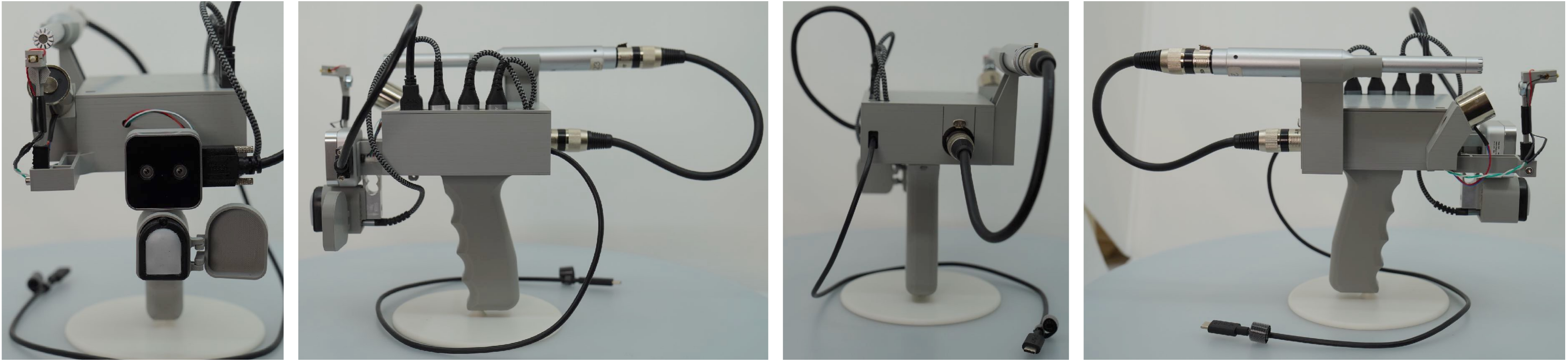}
    \caption{Different views of the \name device. (\textbf{From left}) Front, left, back, and right side view.}
    \label{fig:device_views}
    \vspace{-0.1in}
\end{figure*}

\begin{figure}[h]
    \centering
    \includegraphics[width=\columnwidth]{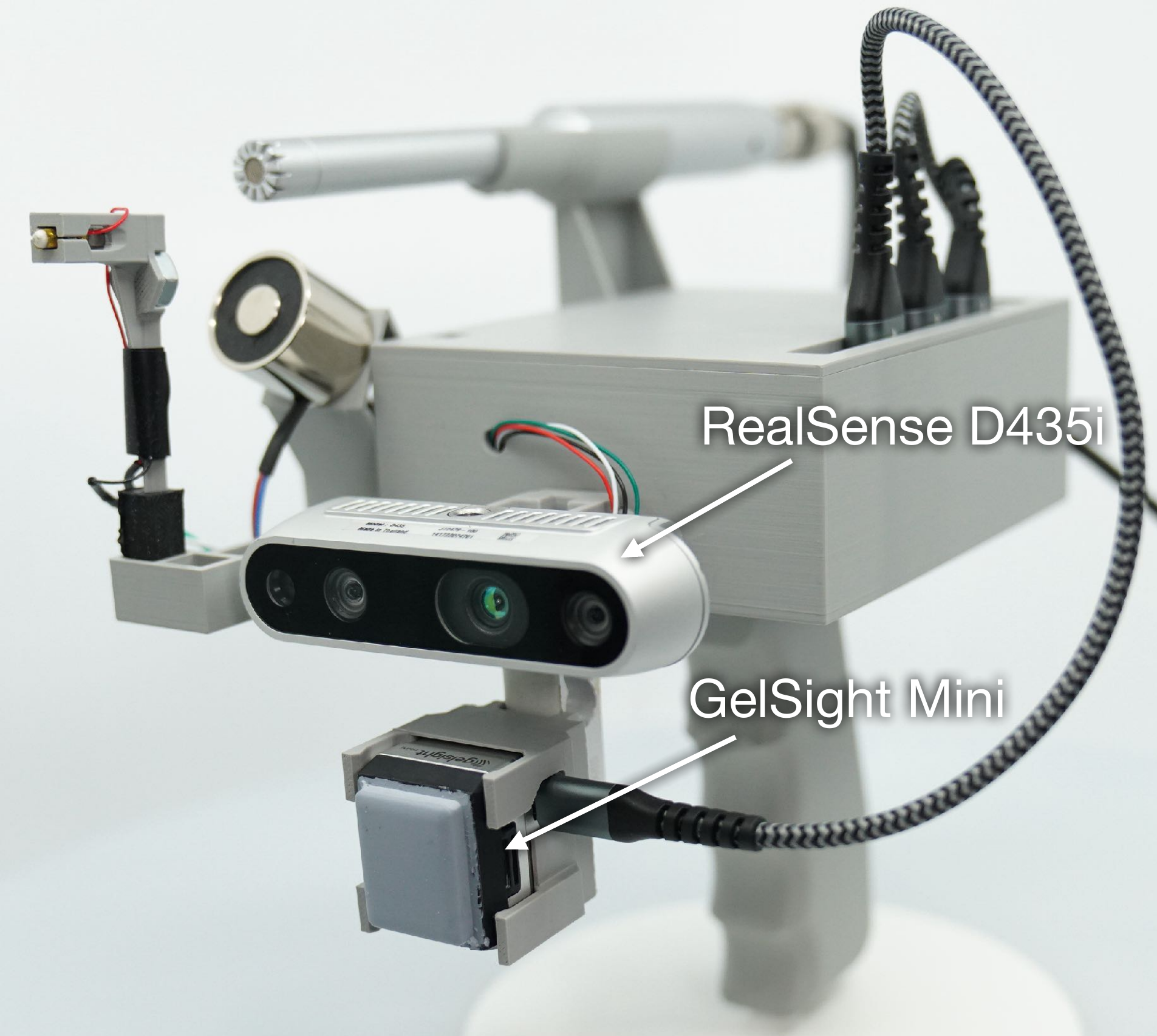}
    \caption{The \name device in a supported alternative configuration, with a RealSense D435 RGBD camera for vision and a GelSight Mini for tactile sensing. The device supports a choice of RealSense D405, D415, D435, or D435i for camera and the DIGIT or GelSight Mini for tactile sensor.}
    \label{fig:alternative_sensors}
    \vspace{-0.01in}
\end{figure}

\begin{table*}[t!]
\centering
\setlength{\tabcolsep}{5pt}
\begin{tabular}{lccc}
\toprule 

\textbf{Part} & \textbf{Unit Cost (USD)} & \textbf{Quantity} & \textbf{Total Cost (USD)} \\
\midrule
\textbf{Sensors}\\
\cmidrule(lr){1-1}
RealSense D405 & 272 & 1 & 272\\
DIGIT & 350 & 1 & 350\\
LIS3DH Accelerometer & 4.95 & 1 & 4.95\\
10kg Load Cell & 12.95 & 1 & 12.95\\
Piezo Stack & 79 & 1 & 79\\
Dayton Audio EMM6 Measurement Microphone & 54.98 & 1 & 54.98 \\
\cmidrule(lr){1-1}
\textbf{Cables} &  &  & \\
\cmidrule(lr){1-1}
1ft MicroUSB & 2.66 & 3 & 7.99\\
USB Micro B with screws & 7.99 & 1 & 7.99\\
4-port USB C hub, 2ft & 9.99 & 1 & 9.99\\
1ft XLR Cable & 6.49 & 1 & 6.49\\
100mm JST connector & 0.75 & 2 & 1.5\\
50mm Qwiic Cable & 0.95 & 1 & 0.95\\
Qwiic Breadboard Jumper & 1.6 & 1 & 1.6\\
XLR Female Panel Mount & 7.63 & 1 & 7.63\\
Female/Male Jumper Cables, 6 inch & 1.95 & 1 & 1.95\\
\cmidrule(lr){1-1}
\textbf{Breakout Boards} &  &  & \\
\cmidrule(lr){1-1}
Raspberry Pi Zero 2W & 15 & 1 & 15\\
Raspberry Pi Headers & 1.05 & 1 & 1.05\\
HiFiBerry DAC2 ADC Pro & 74.9 & 1 & 74.9\\
Qwiic Scale NAU7802 & 16.5 & 1 & 16.5\\
PermaProto Bonnet & 4.5 & 1 & 4.5\\
\cmidrule(lr){1-1}
\textbf{Custom Electronics} &  &  & \\
\cmidrule(lr){1-1}
Aluminum electrolytic capacitor, 33uF 50V 20\%
Resistor, 1k Ohm & 1.01 & 2 & 2.02\\
Resistor, 100k Ohm & 0.24 & 3 & 0.72\\
Resistor, 150k Ohm & 0.24 & 3 & 0.72\\
Tantalum capacitor, 33uF & 2.51 & 2 & 5.02\\
Ceramic capacitor, 3.3uF & 0.5 & 1 & 0.5\\
JST connection header & 0.14 & 2 & 0.28\\
Resistor, 680 Ohm & 0.1 & 3 & 0.3\\
N-channel MOSFET & 1.84 & 1 & 1.84\\
Heat Sink & 0.5 & 1 & 0.5\\
Custom PCB & 3.64 & 1 & 3.64\\
Resistor, 1M Ohm & 0.1 & 1 & 0.1\\
\cmidrule(lr){1-1}
\textbf{Chassis} &  &  & \\
\cmidrule(lr){1-1}
Bambu Basic PLA, Gray 1kg & 19.99 & 0.8 & 15.99\\
Ninjatek Edge Filament, Black 0.5kg & 56.29 & 0.1 & 5.63\\
Various ISO Metric Fasteners & 0.21 & 7 & 1.47\\
\midrule
 \textbf{TOTAL} &  &  & \textbf{971.27}\\

\bottomrule
\end{tabular}
\pretablecaption
\caption{Full bill of materials for building the \name device. Prices do not include shipping costs or taxes.}

\label{tab:bill_of_materials}
\end{table*}

Of all these datasets, those of ObjectFolder are most similar to ours in covering all four sensory modalities and correlating them at a point level. While ObjectFolder 2.0~\cite{gao2022objectfolder} includes more objects, all objects are \textit{virtual} and all sensory readings are \textit{simulated} from these virtual objects. ObjectFolder Real~\cite{gao2023objectfolder} has 100 \textit{real} objects, but sensory readings from each modality are collected in \textit{controlled} environments: the authors collect audio from objects suspended in a semi-anechoic chamber, tactile readings from objects rigidly fixed to a robot table top, and RGB images from objects on a turn-table inside a light-box. Note that even for these 100 real objects, the \textit{point-correlated} RGB and depth readings are \textit{simulated} as well, using renderings of the textured 3D model from a 3D scan. Finally, extending ObjectFolder beyond the 100 real objects required purchasing at least \$11,000 of equipment which must be powered by a wall socket or generator. \name is powered by a laptop and collects additional data at a similar fidelity to ObjectFolder \textit{in the wild} for \$1,000.

\paragraph{Prior Multi-Sensory Data Collection Devices}
We compare the \name device to relevant data collection devices, focusing on devices which lend themselves to capturing object-centric data of one or more modalities in addition to vision, in Table~\appref{tab:device_comparison} of Section~\appref{sec:related}. The UBC ACME~\cite{pai2001scanning} and RealImpact setup~\cite{clarke2023realimpact} both used large, stationary setups where objects were placed in a central position for scanning. Neither were designed to be portable for scanning objects \textit{in situ}. ObjectFolder Real~\cite{gao2023objectfolder} used separate stationary setups for each modality which were also not portable. Most comparable to our device in terms of portability and cost is the novel device introduced with the TVL dataset~\cite{fu2024a}. The device includes a Logitech webcam and a DIGIT sensor fixed to the same chassis such that the webcam is pointed at an oblique angle toward the DIGIT's contact area. This allows for strict temporal alignment between the webcam video and the DIGIT images, but at the cost of the DIGIT occluding contact area from the webcam during contact. While the webcam is also capable of recording ambient audio during contact with the DIGIT, there is no provision for estimating objects' impulse responses at various points by measuring the input-output relation between a precise impact and the sound thereof with the calibrated microphone of our setup.

\section{Additional Details on the \name Device Hardware}
\phantomsection
\label{app:hardware}
The device's enclosure contains many important components including a Raspberry Pi, a HiFiBerry DAC2 ADC Pro, a custom PCB, a USB hub, a voltage converter, an amplifier for the load cell, and an accelerometer. We show photos of the inside of the enclosure in Figure~\ref{fig:device_internals}.
We also include photos of the \name device's exterior from different views in Figure~\ref{fig:device_views}. 

The \name device can be replicated with consumer-grade tools, a 3D printer, and a soldering iron. We include a full bill of materials in Table~\ref{tab:bill_of_materials}, showing the total cost of parts is less than \$1000 (not including shipping costs or taxes).

\myparagraph{Alternative Sensor Support}

We specifically chose the sensors we used on the \name device for their properties that are well-suited to collecting object-centric data in the wild. However, to support additional applications of multi-sensory data capture, we provide designs of mounts for alternative sensors. For vision, we also provide a mount design which supports the RealSense D415, D435, and D435i RGBD cameras, which each have longer minimum and maximum depth ranges than the D405 camera we chose for our dataset and may be especially well-suited for collecting larger objects or scene-centric datasets. For tactile sensing, we provide a mount design which supports the GelSight Mini vision-based tactile sensor, similar to the tactile sensor used in ObjectFolder Real~\cite{gao2023objectfolder}. The GelSight Mini produces high-quality tactile images, at slightly higher cost and lower durability than the DIGIT~\cite{lambeta2020digit}.  We show the \name device with both the RealSense 435 and the GelSight Mini mounted on it in Figure~\ref{fig:alternative_sensors}.

\newpage
\section{Additional Details on the \name Dataset}
\subsection{Objects and Environments}
\phantomsection
\label{app:dataset_details}

\begin{figure*}[htbp]
    \centering
    \includegraphics[width=\textwidth, height=0.92\textheight, keepaspectratio]{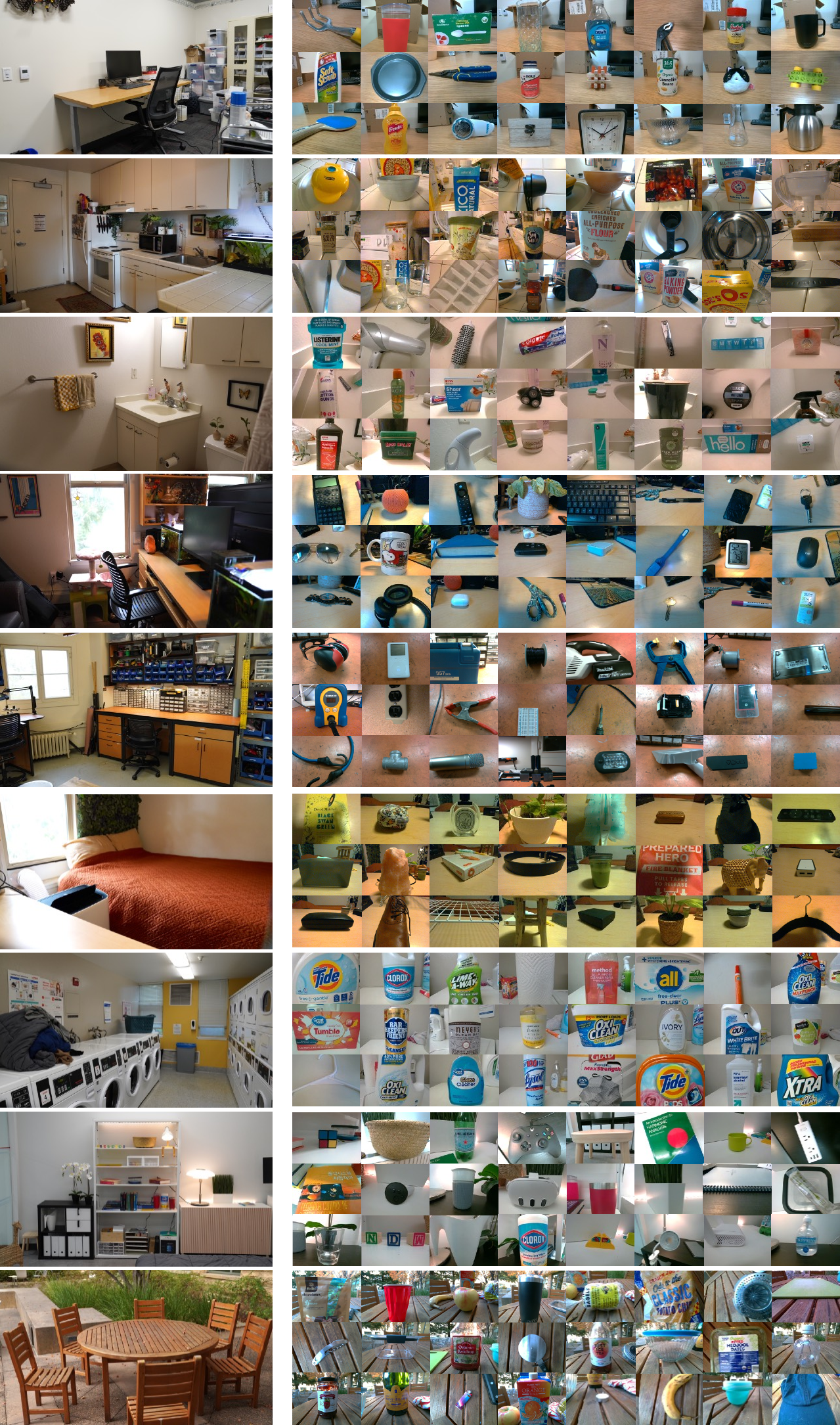}
    \caption{The \name dataset is captured across nine diverse in-the-wild environments. (\textbf{Left}) Photos of the different environments in which data was collected. From top: indoor workspace, kitchen, bathroom, home office, workshop, bedroom, laundry room, living room, and picnic table. (\textbf{Right}) Montages each show 24 distinct objects from each corresponding environment in the left column of the same row. All images in the montages are object images directly lifted from the \name dataset, as captured by the device.  \postcaption}
    \label{fig:envsphoto}
    \vspace{-0.1in}
\end{figure*}

The \name Dataset features a diverse range of objects and environments, which we detail below. We show photos as well as example object images from each environment in Figure~\ref{fig:envsphoto}.

The \textit{indoor workspace} features consistent artificial lighting and minimal noise. Objects consist of diverse materials such as glass, plastic, metal, and ceramic, with varying textures and geometries.

The \textit{kitchen} environment has mixed natural and artificial lighting, creating moderate shadows and highlights. Ambient noises include faint sounds such as a humming refrigerator or occasional outside noise. Objects are primarily food-related, such as packaging, utensils, and glassware, made from cardboard, metal, glass, and plastic.

The \textit{bathroom} environment features artificial lighting with moderate shadows. The matte ceramic sink countertop reduces reflections, and the enclosed space causes slight reverberation. Objects include personal care products with smooth, cylindrical shapes, made from glass and plastic.

The \textit{home office} environment is lit by natural light from windows, supplemented by artificial light. Objects include technology devices (\eg, headphones, remotes), stationery, and decorative items. Audio occasionally includes aquarium bubbling or outdoor noises.

The \textit{workshop} environment is brightly lit with overhead lighting. Objects include tools, hardware, and electronics, made from durable materials like metal and plastic.

The \textit{bedroom} environment has warm artificial lighting that casts strong shadows. Faint sounds from fans or neighboring rooms may be present. Objects include books, plants, and clothing accessories, made from fabric, glass, plastic, and wood.

The \textit{laundry room} features bright, uniform artificial lighting that minimizes shadows. Objects are predominantly cleaning supplies such as detergents, sprays, and bleaches, in smooth plastic containers with vivid packaging. Ambient noises include faint mechanical sounds, but data was collected with laundry machines turned off.

The \textit{living room} environment has soft artificial lighting from overhead and accent lights. Surfaces include metal shelves and wooden tables, with objects like decor, books, plants, and electronics.

The \textit{picnic table} environment features bright, diffuse outdoor lighting. Objects include food items, food storage containers, cutlery, and outdoor accessories, made from plastic, glass, metal, and organic textures (\eg, fruit skins). Ambient noises include occasional distant activity or building hums.

\subsection{Postprocessing}
\phantomsection
\label{app:dataset_postprocessing}
While our RGB, depth, and touch data can be used in their raw state for many useful learning tasks, we postprocess the audio data to normalize differences between recordings with respect to their volume gain settings and the forces of the hammer impacts. We also use the RGB and depth data to generate object point clouds for some experiments.

\myparagraph{Normalizing Audio}
During data collection, the \name device dynamically adjusts the recording gains of both the microphone and the impact hammer for each recording to ensure high signal while also preventing clipping. After collection, we use the annotated gains for each recording in order to scale all recordings to a common gain. For characterizing the input-output relationship of striking the objects, many works deconvolve impact hammer signals from microphone recordings to estimate objects' impulse responses~\cite{rau2019improved, clarke2023realimpact}. However, these works often record more rigid and homogeneous objects with a rigidly positioned impact hammer rig. We found that our hammer signals were too complex to lend themselves to deconvolution without producing filtered noise artifacts. This may be because we record soft, heterogeneous, and articulated objects with a \textit{hand-held} impact hammer rig. Thus, in order to correct for differences in strike force among our audio samples, we simply divide our normalized audio recordings by the peak of their corresponding gain-normalized hammer signals.

\myparagraph{Point Cloud Extraction}

Though the depth readings from the RealSense D405 can be somewhat sparse and noisy depending on the object and capture conditions, we estimate a coherent object point cloud from each captured depth image using the following steps. First, we use DepthAnythingV2 (DAV2)~\cite{yang2024depth} on the captured RGB image to estimate a smooth, dense depth map. Since this predicted depth map lacks a sense of scale, we use least squares regression to estimate a scale and offset which best aligns the prediction map with our sparse real depth recording. To segment the target object from the background and other objects in the scene, we use the Segment Anything Model (SAM)~\cite{kirillov2023segment} on the overlaid RGB image. Since we collect each RGBD image with the center point on some region of the target object, we query SAM with a small disk of points around the center pixel. SAM provides three mask proposals, and we select the mask with the largest area where the center pixel is activated, to favor selecting an entire composite target object rather than an individual section. We apply an erosion to the SAM mask to eliminate ambiguous outlier points. We then apply this final segmentation mask to the adjusted depth map prediction to construct the final object point cloud.

\section{Additional Experimental Results and Examples}
\phantomsection
\label{app:additional_results}

\begin{figure*}[htbp]
    \centering
    \includegraphics[width=\textwidth]{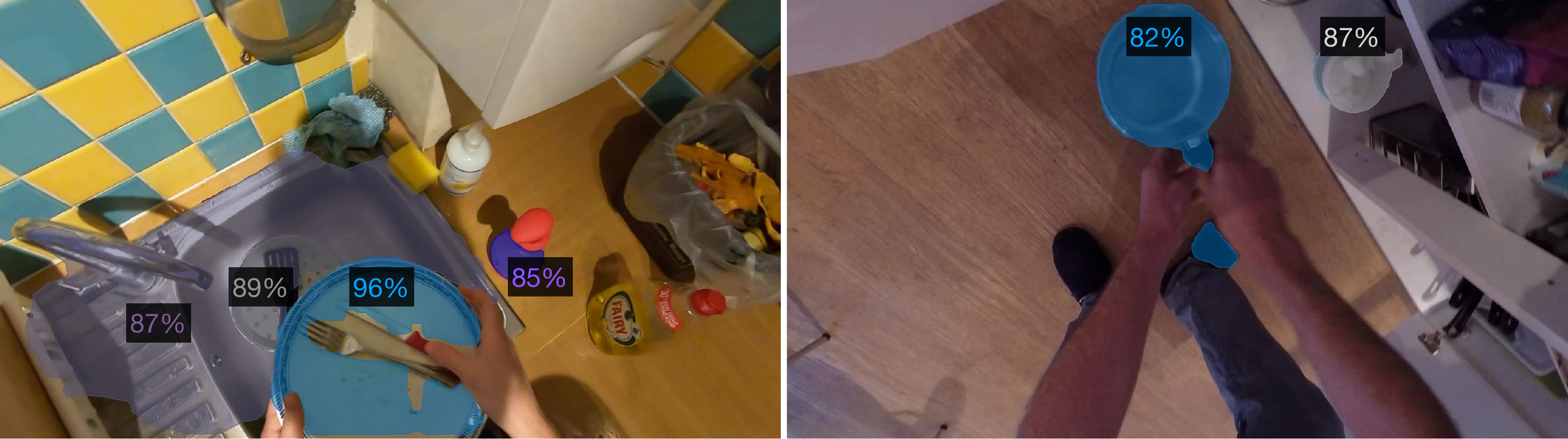}
    \caption{Additional detection results from prompting the Detic~\cite{zhou2022detecting} CLIP-based detector with our audio embeddings of natural impact sounds from egocentric vidoes from kitchens~\cite{Damen2022RESCALING}. (\textbf{Left}) From the sound of setting the red-handled knife on the plate, the detector successfully predicts the correct plate (highlighted blue, bottom center) with highest confidence. It also predicts the other plate, sink, and body of the handsoap dispenser with relatively high confidences, but ignores other objects. (\textbf{Right}) As a failure case, from the sound of the metal pan stacking onto the metal pot below it, the detector predicts the ceramic bowl in the cabinet (highlighted gray at upper right of the image) with highest confidence. It also predicts the correct metal pan (highlighted blue, upper center) with relatively high confidence.}
    \label{fig:additional_detection}
    \vspace{-0.1in}
\end{figure*}

\subsection{Cross-Sensory Retrieval}
\phantomsection
\label{app:additional_retrieval}
\begin{table*}[t!]
\centering
\setlength{\tabcolsep}{5pt}
\resizebox{1.0\textwidth}{!}{
\begin{tabular}{lccccccccccccc}
\toprule 

RGB-Audio-Tactile-Pointcloud  & \multicolumn{3}{c}{RGB$\rightarrow$} & \multicolumn{3}{c}{Audio$\rightarrow$} & \multicolumn{3}{c}{Tactile$\rightarrow$} & \multicolumn{3}{c}{Point$\rightarrow$} &  \multirow{2}{*}{Average} \\

\cmidrule(lr){2-4}\cmidrule(lr){5-7}\cmidrule(lr){8-10}\cmidrule(lr){11-13}

Top-5 Accuracy (\%) & Audio & Tactile & Point &  RGB & Tactile & Point &  RGB & Audio & Point & RGB & Audio & Tactile  \\
\midrule
Random Guess & 5.0 & 5.0 & 5.0 & 5.0 & 5.0 & 5.0 & 5.0 & 5.0 &5.0 &5.0 &5.0 &5.0 & 5.0\\
Out-of-the-Box Pretrained  & 6.0 & 5.4 & 5.4 & 4.8 & 5.2 & 5.4 & 5.8 & 6.0 & 4.8 & 6.2 & 5.6 & 5.4 & 5.5 \\

Fine-Tune w/ Image Loss & 34.0 & 23.3 & \textbf{52.7} & \textbf{40.2} & 12.5 & 17.5 & 23.5 & \textbf{14.4} & 11.9 & \textbf{57.9} & 15.6 & \textbf{14.8} & 26.5\\
Fine-Tune w/ Cross-Sensory Loss & \textbf{38.3} & \textbf{24.4} & 49.0 & 38.8 & \textbf{12.7} & \textbf{20.6} & \textbf{24.4} & 12.1 & \textbf{15.2} & 55.8 & \textbf{18.5} & \textbf{14.8} & \textbf{27.0}\\

\bottomrule
\end{tabular}
}
\pretablecaption
\caption{Cross-sensory retrieval top-5 accuracies of the cross-sensory encoder ensemble trained with different strategies using our dataset. The top and bottom column headers denote the query and retrieved modalities, respectively. The encoders' Out-of-the-Box weights do not generalize well to our data across modalities, though Fine-Tuning performance of each loss type varies by modality.}

\label{tab:cross_modal}
\end{table*}

We show additional results from the experiment described in Section~\appref{sec:cross-sensory_retrieval} for the multi-modal encoder ensemble in Table~\ref{tab:cross_modal}.

\subsection{Cross-Sensory Contact Point Localization}
\phantomsection
\label{app:additional_localization}
\begin{table*}[t!]
\centering
\setlength{\tabcolsep}{5pt}
\resizebox{1.0\textwidth}{!}{
\begin{tabular}{lccccccccccccc}
\toprule 

RGB-Audio-Tactile-Pointcloud  & \multicolumn{3}{c}{RGB$\rightarrow$} & \multicolumn{3}{c}{Audio$\rightarrow$} & \multicolumn{3}{c}{Tactile$\rightarrow$} & \multicolumn{3}{c}{Point$\rightarrow$} &  \multirow{2}{*}{Average} \\

\cmidrule(lr){2-4}\cmidrule(lr){5-7}\cmidrule(lr){8-10}\cmidrule(lr){11-13}

Top-1 Accuracy (\%) & Audio & Tactile & Point &  RGB & Tactile & Point &  RGB & Audio & Point & RGB & Audio & Tactile  \\
\midrule
Random Guess &16.7 &16.7&16.7&16.7&16.7&16.7&16.7&16.7&16.7&16.7&16.7&16.7&16.7\\
Out-of-the-Box Pretrained  & 16.0 & 17.2 & 16.7 & 16.7 & 16.7 & 16.8 & 16.7 & 13.9 & 16.0 & 15.8 & 17.7 & 15.8 & 16.3\\
Fine-Tune w/ Image Loss & 19.8 & 19.3 & \textbf{35.1} & 19.4 & \textbf{20.7} & 19.6 & \textbf{24.3} & \textbf{19.4} & 20.5 & \textbf{44.4} & 19.7 & 20.3 & 23.5\\
Fine-Tune w/ Cross-Sensory Loss & \textbf{20.7} & \textbf{23.6} & 31.5 & \textbf{22.2} & 18.2 & \textbf{21.5} & 22.9 & 17.1 & \textbf{25.6} & 41.2 & \textbf{22.0} & \textbf{24.7} & \textbf{24.3}\\

\bottomrule
\end{tabular}
}
\pretablecaption
\caption{Contact localization top-1 accuracies of the cross-sensory encoder ensemble trained with different strategies using our dataset. The top and bottom column headers denote the query and retrieved modalities, respectively. Their Out-of-the-Box weights generalize poorly to our data across all modalities, but the highest performance loss for Fine-Tuning varies by modality.}

\label{tab:cross_modal_intra}
\end{table*}

We show additional results from the experiment described in Section~\appref{sec:cross-sensory_contact} for the multi-modal encoder ensemble in Table~\ref{tab:cross_modal_intra}.

\subsection{Zero-Shot Audio-Based Object Detection}
\phantomsection
\label{app:additional_detection}
We show additional results from the experiment described in Section~\appref{sec:audio_detection} in Figure~\ref{fig:additional_detection}.

\subsection{Training and Testing Details}
\phantomsection
\label{app:training_details}
During training for each experiment, we augment the image modality using strong image augmentations as in MoCo~\cite{he2020momentum}. We use all other modalities as is, without augmentation.
We train with a batch size of 64, using the AdamW optimizer~\cite{loshchilov2018decoupled} with a learning rate of $10^{-5}$. For all experiments except those in Section~\appref{sec:objectfolder_pretraining}, we train for 500 epochs. For the experiments in Section~\appref{sec:objectfolder_pretraining}, to avoid overfitting on the small fine-tuning set from ObjectFolder Real, we train for only 50 epochs on each dataset.  During evaluations for all experiments, we evaluate each model on the entire test set with five different random samplings of object points and report the average to reduce variance.

\end{document}